# Deep Learning-based Sentiment Classification: A Comparative Survey

**MOHAMED KAYED[1], REBECA P. DÍAZ REDONDO[2], AND ALHASSAN MABROUK[3]**

[1]Computer Science Department, Faculty of Computers and Artificial Intelligence, Beni-Suef University, Egypt
[2]Information & Computing Lab, AtlanTTIC Research Center, Telecommunication Engineering School. Universidade de Vigo, Spain
[3]Math. and Computer Science Department, Faculty of Science, Beni-Suef University, Egypt

Corresponding author: Mohammed Kayed (mskayed@gmail.com).

**ABSTRACT** Recently, Deep Learning (DL) approaches have been applied to solve the Sentiment Classification (SC) problem, which is a core task in reviews mining or Sentiment Analysis (SA). The performances of these approaches are affected by different factors. This paper addresses these factors and classifies them into three categories: data preparation based factors, feature representation based factors and the classification techniques based factors. The paper is a comprehensive literature-based survey that compares the performance of more than 100 DL-based SC approaches by using 21 public datasets of reviews given by customers within three specific application domains (products, movies and restaurants). These 21 datasets have different characteristics (balanced/imbalanced, size, etc.) to give a global vision for our study. The comparison explains how the proposed factors quantitatively affect the performance of the studied DL-based SC approaches.

## I. INTRODUCTION

Websites usually provide services for different purposes such as booking services [1], social media communication [2], political, E-Commerce (EC) and blogging. These sites allow users to share their viewpoints/opinions in forums [3], blogs [3], [4], reviews [1], news articles [3], [5], [6] or wikis. Consequently, they are considered as one of the most important sources of consumers (users) opinions. Within this context, a new research area, Sentiment Analysis (SA), has emerged with the aim of analyzing and categorizing user opinions. Consequently, SA is a relevant source of information for different application fields, such as sales measurement, political movements or prediction of election results. In fact, both individuals and organizations are really interested in SA. For instance, in the EC field, customers of big companies such as Amazon, Walmart or eBay (among others) seek opinions from other users before making a purchase. Additionally, companies carefully check and process these opinions (complaints or positive reviews) to enhance their products and to take strategic decisions for stocks and, even, production. However, these opinions are written in natural language, using colloquial forms, jargon and including abbreviations, lack of capitals, problems with spelling, punctuation and grammar errors. Moreover, texts usually include other elements such as URLs, tags, and other unstructured data. This makes it easier for users to read, but more difficult for machines to process; so, it is a challenge that requires a combination of different techniques to be managed. In fact, customer reviews must be pre-processed in order to be converted into structure data before being analyzed.

In review mining, many approaches have been proposed [7] with different SA tasks: aspect extraction [8], [9], Sentiment Classification (SC), ambiguous text [6], subjectivity classification [10] or opinion spam [11]. These approaches have worked on different domains [12], [13] or languages [3], [14], [15]. Among all of these approaches, SC is considered as one of the key tasks. In SC, the main goal is inferring the polarity of a given message or review [4], [16], [17]. That is, the text, document, sentence or feature is analyzed in order to know if it is positive, negative or neutral. Some websites allow users to directly evaluate items or services using stars [3], numbers or a thumb up/down [18], [19]. However, customers also have the opportunity to write his/her opinion to supplement the assessment. Therefore, even in this case, it is still necessary to extract knowledge and polarity from customer opinions in order to identify ambiguous text, detect opinion spam or infer knowledge for strategic decisions.

In the specialized literature, there are different proposals for Sentiment Classification (SC) of textual reviews. These approaches are usually classified into three types [20]: Lexicon [21], Machine Learning (ML) [4] and Hybrid-based approaches [22]. Lexicon-based are fast in training, but ML-based ones achieve state-of-the-art performance in SC. Hybrid-based proposals are characterized by its high complexity, so they are not popular yet. Consequently, ML-based approaches are, without doubt, the most popular for SC with models that have been repeatedly used in the literature such as Naïve Bayes, Maximum entropy, Decision trees, Support Vector Machines or Neural Networks (NN). In fact, the last ones, NN, are widely used [23], [24], [25], [26] because of its higher efficiency (high performance and fast execution) compared to the other alternatives.

Within the NN field, Deep Learning (DL) approaches have made a breakthrough in SA [25], so different researchers have analyzed DL-based SA proposals. In [26], the authors compare the performance of different DL methods for different SA tasks. However, in [27], the authors focus only on one of them: a deep comparison of the performance of aspect extraction.

To the best of our knowledge, there is no a prior comparative study of performance about the application of DL in Sentiment Classification, although many approaches have been proposed on different levels of SC: aspect [28], [29]; sentence [30], [31]; and document [32], [33]). Therefore, this paper provides a comparative and a comprehensive literature-based survey for the existing DL approaches that are suggested to solve the task of SC. More specifically, our study targets proposals within the field of DL-based SC for customer reviews.

Consequently, the paper has two main contributions. First, it discusses the utmost key factors that affect the performance of DL-based SC approaches. These factors are classified into three different categories: data preparation based factors, feature representation based factors and the techniques based factors. Second, we compare the performance of more than 100 DL-based approaches on SC that have been published in the specialized literature. These approaches have been assessed on 21 widespread datasets of three specific review domains (products, movies and restaurants). All datasets have different characteristics (balanced/imbalanced, size, etc.), which gives a global vision in our study.

The remainder of the paper is organized as follows. Section II shortly discusses a background knowledge about neural network and DL. Section III discusses the factors that affect the performance of DL-based SC approaches. The set-up of the experiments are reported in Section IV. The results of recent DL-based SC approaches on different domains are presented and compared in Section V. Extra factors that could be used to enhance the performance of DL-based SC approaches as well as open issues on SC are presented in Section VI. Finally, Section VII concludes our work and expounds our proposal for the future work.

## II. NEURAL NETWORK

Neural Networks (NNs) have different advantages: (i) NNs provide a nonlinear model, which is flexible in representing complex relationships; (ii) NNs are able to estimate the posterior probabilities (i.e., performing statistical analysis); (iii) the execution time of an NN model is not excessively long; and (iv) they give good performance even with noisy data. Therefore, NNs have been widely used in sentiment analysis. In this section, we provide a brief introduction about NNs and DL, and how to train DL models.

### A. NN Architecture

A NN is a network diagram that interconnects nodes, also known as neurons, which transmit data among them and are arranged in layers, as Figure 1 shows. Each neuron in the network has a certain random weight. Also, each layer has a bias that influences the output. A neuron is a simple mathematical model that calculates an output value in two steps [4]. First, it calculates an input function as the sum of weights and the values of the input neurons. Second, it applies an activation function to this sum to provide the output. The activation function (e.g., Sigmoid, Tanh or Relu) is typically a nonlinear function. Relu activation function is the easiest to compute, the fastest to converge in training and yields equal or better performance in NNs [34].

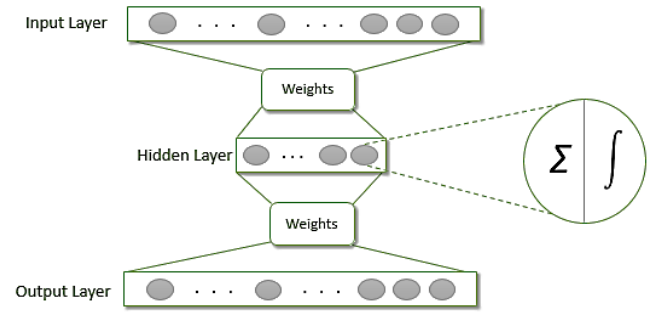

**FIGURE 1.** Neural network architecture.

The architecture of NN includes three layers: input, hidden and output layer. The input layer is a *word embedding vector*, which will be discussed later. Secondly, the hidden layer takes the input features and gives an output to the next layer using the previously mentioned activation function. Finally, for each class in the output layer, there is a probability distribution via a softmax function.

### B. Deep Learning: A Brief Overview

Despite of the previously mentioned advantages of NNs, there is a clear shortcoming for SC applications. When the number of input and hidden nodes increases, an overfitting problem occurs due to the increase of parameters within the NN. To address this problem, deep learning (DL) methods reduce the number of parameters, but they include multiple layers. The connection weights among neurons can be adjusted to perform tasks such as classification with the aim of achieving high performance in SC [35], [36], for instance.

DL models are trained, i.e. the weights among neurons are adjusted, by using a backpropagation process [37] in which a loss function is minimized using the Stochastic Gradient Descent (SGD) algorithm. Gradients of the loss function calculate weights from the last hidden layer to the output layer. Then, these weights are recursively calculated by applying the chain rule backwards. SGD is an iterative refinement process that stops when a certain stopping criteria is met. It estimates the parameters for each training sample as opposed to the whole set of training samples in batch gradient descent. For example, the training values are loaded into the input nodes. If misclassification occurs, the error is propagated back through the network, modifying the weights to minimize the error.

It is necessary to remark that DL models have a high computational cost, such as high memory usage, for instance. However, DL models provide very good results at SA tasks [38], [39]. In fact, different DL models have been used to solve the SC problem, including Convolutional [40], Recurrent [32], [33], [41], Recursive [42], [43] and Hybrid Neural Networks [25], [44].

## III. PERFORMANCE OF DL-BASED SC APPROACHES

This section addresses the factors that affect the performance of DL-based SC approaches. As shown in Figure 2, these factors are classified into three different categories: data preparation based factors, feature representation based factors and classification techniques based factors. The next three subsections will discuss these three categories in detail.

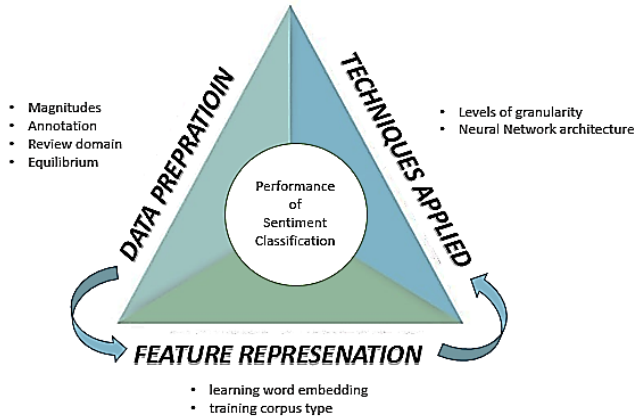

**FIGURE 2.** Factors that affect the performance of DL-based SC approaches.

### A. DATA PREPARATION

Data or Corpora preparation is one of the fundamental challenges to generate useful information. In our case, the process of data preparation always starts by extracting users' reviews from the websites, handling the gaps in these reviews and converting them into a useable format such as the one in Figure 3.

Corpora are usually classified into three different types: polarity, subjectivity and ironic [6], [12]. A polarity corpus classifies the reviews into a specific number of categories (positive, negative, etc.). A subjective corpus extracts opinions from reviews or texts; for instance, Pang and Lee [10] used a cut-based classification to determine the subjectivity. Finally, an ironic corpus, the most complex in this field, extracts irony from the reviews [5], [6], [45]. Our contribution only focuses on the first one, Polarity or Sentiment Corpora, but expressed at different levels: documents [16], [46], [47], [48], sentences [3], [17] or aspects [49], [50]. The last one is commonly used in SC.

We suggest that four factors could affect the performance of SC during the polarity corpus preparation (see Figure 2):

**a) Magnitudes**: the number of classes may be binary, ternary or multiple (scale) classes. The first one, *Binary corpora* ("positive" and "negative"), ignores the neutral class. Some approaches, like [4], [17], [23], [22], [51], directly consider the vagueness (neutrality) as noise. Others, like in [16] consider them as positive opinions, since there is no negativity in the review. The second one, *Ternary corpora*, splits the reviews into three classes: "positive", "negative" and "neutral" [52]. Finally, *Scale corpora*, matches the reviews with a number, following a scale similar to the star ratings in websites, like in [54].

**b) Annotation:** Data is usually divided into two groups: labeled data and unlabeled data. The former entails the magnitude is known (the scale positive, negative, etc.). The later entails this factor is unknown. For instance, in [51], labeled data is split into three classes (negative, positive and neutral).

**c) Review domain:** In our study, we focus on three application fields or review domains: products, movies and restaurants. In all of them, we analyze the behavior at the three considered levels: document-level [4], [10], [16], [17], [18], [23], [53], [54], [55]; sentence-level [50], and aspect-level [5], [56]; which is explained in Figure 3.

**d) Equilibrium:** Corpora may be classified as balanced or imbalanced. On the one hand, in balanced corpora, data are partitioned and distributed equally on different classes. For example, in [10], [17] the corpus includes 2,000 movie reviews from *rottentomatoes*, which are divided into two balanced categories (+ve and –ve). In [57], [58], the analysis is done using a multi-domain corpus of 400 reviews from *Epinion* that are distributed on 8 balanced categories of 50 reviews each (music, computers, movies, phones, books, cookware, cars and hotels). On the other hand, imbalanced corpuses have classes with significantly different number of elements. For example, SINAI dataset in [16] includes about 2,000 documents (reviews) of 10 camera models, where 93% were classified as positive. In Dave *et al.* [54] a corpora from CNET and Amazon with 7 imbalanced categories is used. However, Rushdi Saleh *et al.* [16] achieved promising results working with an imbalanced data set of product reviews.

### B. FEATURE REPRESENTATION

DL models are oriented to automatically learn the best representations (features) to return a probability that could be used to assign a given label to each input [59], [60]. Figure 4 compares the architecture of (i) traditional ML-based approaches and (ii) DL-based approaches on SC. It is necessary to remark that in both cases, there is a previous stage, pre-processing, needed to clean and organize the data. After that, the first approach, ML-based models, consists of three consecutive stages: transformation (e.g. Term Frequency—Inverse Document Frequency), feature selection (e.g., chi-square) and the construction of the classifier. Some studies have demonstrated that working on the feature selection stage can potentially improve the classification accuracy [61]. After the pre-processing phase, the second approach, DL-based models, only uses feature representation (word embedding). This entails a more flexible and adaptable solution to the data [62]. Generally speaking, DL-based approaches give better performances than traditional ML-based approaches with large datasets, since DL-based models require large sets of training data.

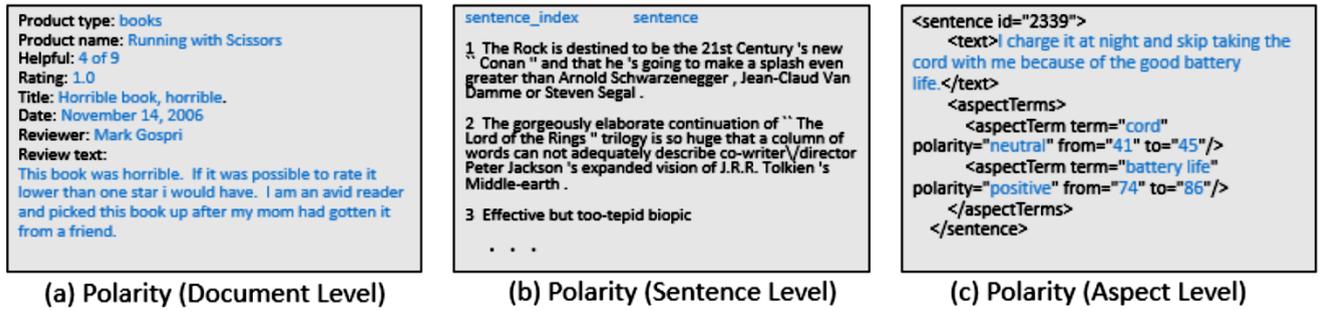

**FIGURE 3.** Tuples output of polarity structure reviews on different levels.

Word embedding, typically applied for feature representation in DL-based approaches, transforms words or phrases into vectors of real numbers, which are also known as word vector representations or distributed representation. In short, word embedding applies a dimensionality reduction that gets a lower-dimensional dense vector space from a high-dimensional sparse embedding. Each dimension of the embedding vector represents a latent feature of a word. This means, it describes syntactic and semantic properties of the words [48], [63], [64].

This strategy was firstly proposed in [37] and it has been widely used in SA [65], [66]. Word embedding handles many SA tasks such as aspect term extraction [67], SC [32], [62], [68], [69], among other. The specialized literature offer different alternatives (algorithms) to generate word embedding. For example, the Sentiment-Specific Word Embedding (SSWE) method is proposed in [70], and its features offer competitive results in context-based SA [63], but it gave a poor performance in Aspect-Based Sentiment Analysis (ABSA) [39]. In [32], the quality of representations is improved for longer documents. Additionally, a Bilingual Sentiment Word Embedding (BSWE) model for cross-language SC is proposed in [71].

We consider that the performance of SC during the feature representation phase can be affected by two factors, as depicted in Figure 2:

**a) Learning word embedding:** This is one of the most important factors that affects the performance of DL-based approaches. There are usually two methods to create word embedding versions: matrix factorization and neural network.

The first approach, Matrix factorization-based word embedding is a linear algebra process that applies rank reduction on a large term-frequency matrix. This matrix measures co-occurrence of terms in two frequencies: the rows of Term-Document frequencies matrix represent words and the columns represent the documents or the paragraphs. On the other side, both rows and columns of the Term-Term frequencies matrix represent words. In [72], the authors propose applying word embedding with matrix factorization for personalized review-based rating prediction. Besides, there are many techniques that are based on observed co-occurrence patterns to reduce the dimensionality such as clustering, Latent Dirichlet Allocation (LDA) and Singular Value Decomposition (SVD) [73].

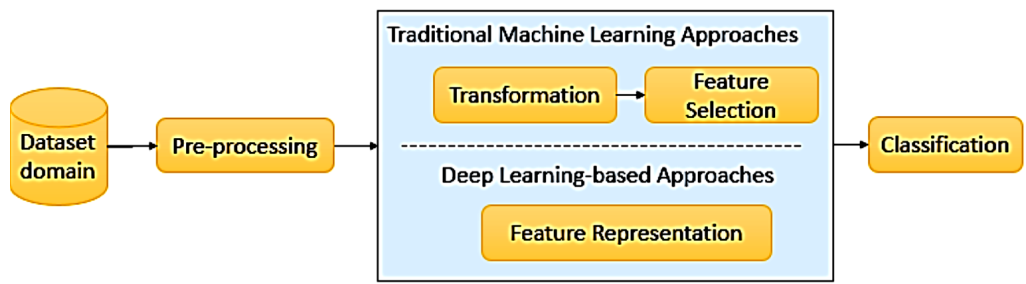

**FIGURE 4.** Comparison between traditional ML and recent DL-Based approaches for SC.

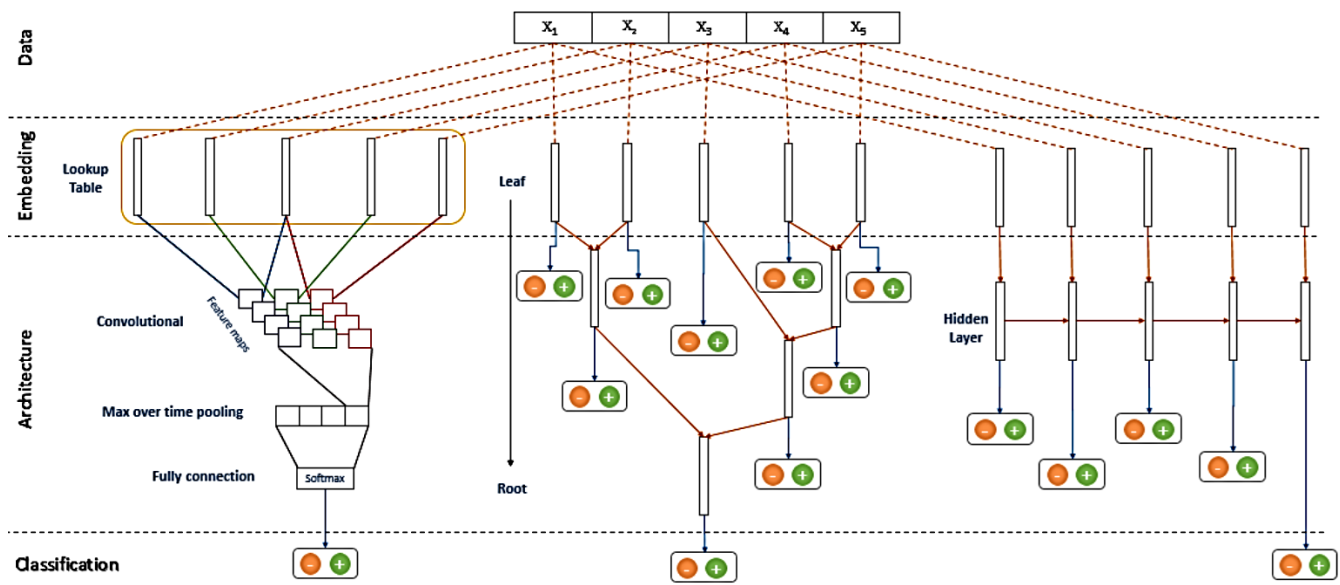

**FIGURE 5.** Convolutional neural network (left), recursive neural network (center), and recurrent neural network (right). The architectures for SC.

The second approach, Neural network-based word embedding has words as input and gives context as an output. For example, a neural probabilistic language model (NPLM) [59] is the first word embedding model with a shared lookup table. Given a word and its previous words, the model predicts the probability function for its next word. Wang *et al.* [74] developed a neural architecture to train a sentiment bearing word embedding by integrating the sentiment supervision at both the document and the word levels.

**b) Training corpus type:** The performance of SC is clearly dependent on the corpora used. For instance, in [67], [66] some options are analyzed and the conclusion is that Amazon corpus has a better performance on SC than both Wikipedia and Google news corpora, because Amazon corpus contains more opinion words.

### C. APPLIED CLASSIFICATION TECHNIQUES

Taking into account the techniques applied for SC, we consider that there are two factors that could affect the performance, as shown in Figure 2. On the one hand, the level of granularity and, on the other hand, the neural network architecture used. NNs have emerged as a good approach for SC, so we have focused only on this approach.

**a) Level of Granularity:** Several DL-based approaches have been proposed to solve the SC problem at the three levels: document, sentence and aspect. Table 1 lists the most recent and common approaches on each one.

**TABLE 1.** Allocations of papers based on SC-levels with the four types of NN.

| NN Architecture | Document | Sentence | Aspect |
|---|---|---|---|
| Convolutional | [23], [25], [75], [76] | [38], [40], [45], [62], [77], [78], [79], [80], [81], [82] | [83] |
| Recurrent | [25], [33], [84], [85], [86], [87], [88], [89], [90], [91] | [31], [65], [79], [80], [81], [92], [93], [94], [95], [96], [97], [98] | [99], [28], [29], [35], [39], [100], [101], [102], [103], [104], [105], [106], [107] |
| Recursive | -- | [30], [36], [42], [43], [52], [108], [109], [110], [111], [112], [113] | [49] |
| Hybrid | [5], [25], [55], [114] | [44], [115], [116], [117] | -- |

**b) Neural Network Architecture:** As depicted in Figure 5, a NN works loading the words (X) into the input nodes (like word embedding) and the final values of output nodes give the classification results. Generally, it is accepted that there are four types of NN architectures that try to capture semantic and syntactic information: (i) Recursive Neural Networks (RecNN); (ii) Recurrent Neural Networks (RNN); (iii) Convolutional Neural Networks (CNN); and (iv) hybrid approaches. Table 1 classifies different approaches that can be found in the literature according to the type of NN. Table 2 gives a brief comparison between these approaches summarizing the goal, the advantages and disadvantages of each one. Please, note that hybrid approaches are not included in Table 2 because there are not common characteristics, since they depend on the types of NN that are combining.

**TABLE 2.** Advantages and disadvantages of the most common NN models on SC.

| NN Architecture | Goal | Advantages | Disadvantages |
| --- | --- | --- | --- |
| Convolutional | The essence in CNN are (i) the non-linearity of the model and (ii) the ability to learn embedding for small fixed size regions. | It overcomes the high computational cost of NNs. | It requires more training time compared with other techniques |
| Recurrent | RNN can capture sequential information in flexible computational steps. | It reduces the number of parameters needed to learn. | The output of one state depends on the previous state. Thus, it needs from huge amount of memory. |
| Recursive | RecNN is a generalization of RNN that applies recursively the same set of weights over a directed acyclic, but in a tree structure input. | It elegantly learns compositional semantic in simpler structure. | The application on SC still requires further research, which leads to inaccuracies. |

In parallel, these models can be split into two types: feedforward neural network (e.g., CNN) and backward neural network (e.g., RNN and RecNN). The four NN architectures (CNN, RNN, RecNN and Hybrid neural networks) will be discussed in the next four subsections, including information about the different DL methods.

### 1) CONVOLUTIONAL NEURAL NETWORK (CNN)

CNN has become a popular DL model which is used by many researchers in various domains such as image processing and Natural Language Processing (NLP). In NLP, the words are converted into vectors [40], [62], [75], [78], [82]. CNN is characterized by the non-linearity of the model and its ability to learn embedding from small size regions. Its architecture is shown in Figure 5 (left) and it consists of four layers: an embedding (input) layer, a convolutional layer, a pooling layer and a fully connected (output) layer. In the first one, the embedding layer, each review is embedded at word-level and it is represented as a matrix. In the second one, the convolutional layer, the width of a filter is fixed to the dimensionality of the word vector in order to capture the relationships among adjacent words, producing a feature map. Thus, this layer (convolutional one) captures important features (n-gram) for each feature map (i.e., a specification of a semantic/structural feature) by directly applying word vectors. Both word vectors and the shared (word independent) kernels are the parameters of CNN, which can capture the predictive structures of small regions. In the third layer, pooling layer, the max-over-time pooling operation extracts the maximum value corresponding to each feature map. The pooling scheme can also be used to deal with the variant lengths of the feature maps produced by filters of different sizes. Besides, the max-pooling layer may produce a fixed-length output regardless of the size of the filter window. Last, in the fourth layer (output), the features are extracted and concatenated in the fully connected layer, which has a probability distribution over the output classes. It is worthy to notice that a deep CNN may have more than four layers: for instance, one input layer, two convolution layers, two max-pool layers and a fully connected layer with a softmax output. Experiments show that deep CNN models, even without any feature engineering or linguistic patterns, still outperforms state-of-the-art models. Dos Santos *et al.* [78] focused on deep CNN for sentiment detection in short texts. X. Zhang *et al.* [24] showed that, for text classification, a deep CNN over characters performs properly.

The CNN model is applied in many tasks. Tang, Qin *et al.* [25] used hierarchical structure in SC. Guan *et al.* [82] proposed a novel DL framework for review SC that employs prevalently available ratings as weak supervision signals. Yu *et al.* [62] learned sentence embedding using two auxiliary tasks (whether the sentence contains a positive or a negative domain-independent sentiment word), which did not predict of pivot features on a large set. X. Zhang *et al.* [24] applied a character-level CNN for text classification and achieved competitive results. CNN can extract local n-gram features within texts but may fail to capture long-distance dependency, Long Short Term Memory (LSTM) can address this problem by sequentially modeling texts [79]. Many papers used multiple algorithms providing other advantages. CNN and RNN are often combined with sequence-based or tree-structured models [40], [25], [38], [41]. J. Wang et al. [79] used a regional CNN-LSTM to predict the valence arousal ratings of texts. LSTM was used in combination with CNN to perform SA for short text, which achieved better performance in terms of accuracy in [80]. The experiment showed that CNN is an alternative to overcome the high computational cost of NN, but it requires more training time compared with other techniques.

Below, we summarize some CNN models that have been applied on SC: DCNN [38], CNN [40], MVCNN [118], Seq2-bown-CNN [75], UPNN [119], CNN-Rule-q [120], Char-level CNN [24] and PF-CNN [83]:

**DCNN**: A Dynamic CNN that applies the dynamic k-max pooling strategy (i.e., selecting the k most active features) for sentence modeling.

**CNN**: A model that uses two channels for sentence classification. First, a static channel that is kept static throughout the training. Second, a non-static channel that is fine-tuned for each task. Multichannel applied the two channels for each filter.

**MVCN**: A multichannel variable-size CNN that introduces multichannel embedding for sentence classification.

**Seq2-bown-CNN**: It groups texts information as a sequence. It has two sequential convolution layers and another convolutional layer to represent the entire document with a bag of words.

**UPNN**: It learns the semantic representations of user and product in a bottom-up way.

**PF-CNN**: The authors use the information of aspects on CNN by parameterized filters and parameterized gates.

**3W-CNN**: This method helps to reclassify the predictions by three ways to improve CNN by NBSVM.

**CNN-Rule-q**: This method changes the structural information of the logical rules to weight the NN through an iterative distillation method.

### 2) RECURRENT NEURAL NETWORK (RNN)

RNN has become popular in SC because it can capture sequential information in flexible computational steps. The RNN model has two important features compared to the CNN. First, CNN has different parameters at each layer, while the parameters in RNN are the same at each step (i.e., it reduces the number of parameters needed to learn). Second, in RNN, the output of one stage depends on the previous stage, thus it needs a huge memory. Therefore, RNN is more superior in processing sequential information compared to CNN.

Therefore, RNN is a robust network architecture (Figure 5, right) for processing sequential data. It allows cyclical connections and reuses the weights across different instances of neurons, each of them associated with different time steps. This idea can explicitly support the network to learn the entire history (i.e., current states) of previous states. With this property, RNN maps an arbitrary length sequence to a fixed length vector.

The simple RNN has limitations caused by the gradient, making it difficult during training in the backpropagation process. The two main problems are: (i) the gradient vanishing problem (i.e., the gradient comes close to zero) and (ii) the exploding gradient problem (i.e., being extremely high) which makes the learning process unstable. Tuning the parameters may improve the gradient. This limitation was reduced by the introduction of networks such as Long Short-Term Memory (LSTM) [121] and Gated Recurrent Units (GRU) [122]:

**LSTM** takes the whole document as a single sequence and the average of the hidden states of all words is used as a feature for classification. LSTM cannot extract the aspect information, this is why it achieves less performance. Therefore, many upgrades showed in [28], [103], [29], [39], [86], [123]. LSTM generally outperforms that of the CNN model [87], [76]. Xin Wang *et al.* [65] proposed encoding entire tweets with LSTM, whose hidden state is used for predicting sentiment polarity. Qian *et al.* [95] proposed linguistically regularized LSTMs for SA with sentiment lexicons, negation words and intensity words.

**GRU** is similar to LSTM and it has not a memory unit. J. Cheng *et al.* [124] proposed a bidirectional GRU model to attend the aspect information for one given aspect and extract sentiment for that given aspect.

Additionally, there are extensions of RNN such as Bidirectional Recurrent Neural Networks (BRNN) [125]. BRNN incorporates a forward and a backward layer in order to learn information from preceding and following tokens. Also, Gate Recurrent Neural Networks (GRNN) [25] handles the document level SC, which obtain hierarchical representations by firstly building representations of sentences, and then aggregating those into a document representation. Moreover, in [93], BRNN is combined with LSTM to result in Bidirectional LSTM (BLSTM), which can access the long-range context in all input directions and more structure information. Ruder *et al.* [103] created a hierarchical BLSTM with a sentence level which is used as the input of a review level. Thus, BLSTM allows them to take into account inter- and intra-sentence context. They used only word embedding to make their system less dependent on extensive feature engineering or manual feature creation.

However, RNN has a remarkable problem processing the information (in the traditional encoder-decoder framework) that may entail the encoding of irrelevant information. One possible solution is to employ an *attention mechanism*, which allows the model to learn on which part of the text must focus. As an example, the work in [29] cannot know important words, because it did not use this attention mechanism. Tang, Qin, Feng, *et al.* [39] kept attention on aspect phrases. The general idea of the attention mechanism is to compute an attention weight from each lower level and, then, aggregate the weighted vectors for the higher level representation. This mechanism is suitable to be applied to SC, since it can focus on the important parts of the sentence. In the attention mechanism, the local semantic attention represents the implementation [87] by introducing a hierarchical network with two levels of attention mechanisms for document classification: word attention and sentence attention. Attention-based methods in [88] use attention mechanisms to build representations by scoring input words or sentences differently. Consequently, it is able to learn distributed document representations in Chinese and English. Also, this issue is solved by using external memory such as the *Memory Networks model* (MemNet). Memory networks have played a major role in ABSA task [35], [106], [99]. Recent researches show that memory network obtains the state-of-the-art results in SC [89]. Recurrent attention models have achieved superior performance [107] by learning a deep attention over the single level attention. Thus, multiple passes (or hops) over the input sequence could refine the attended words again and again to find the most important ones. In fact, Tang, Wei, *et al.* [63] adopted a memory network (MemNet) solution based on multiple-hop attention.

RNN models have been widely used in SC, such as those using LSTM (e.g., Cached LSTM [86], TC-LSTM [39], TD-LSTM [29], ATAE-LSTM [28], HP-LSTM [103], and AF-LSTM [123]), those using attention mechanism (e.g., HN-ATT [87] and Structured Att [126]) and those using memory network (e.g., MemNet [35], DMN [102]). Skip-thought [92] and Byte mLSTM [32] used RNN for a word prediction. Other models are NCSL [31], Virtual Adversarial (VIRTUAL ADV) [68] and TopicRNN [33]:

**Cached LSTM**: It improves the LSTM ability to carry information for long distances.

**NCSL**: Neural Context-Sensitive Lexicon utilizes sentiment lexicons to treat the sentiment score of a sentence as a weighted sum of prior sentiment scores of negation words and sentiment words.

**TD-LSTM**: Target-dependent LSTM treats an aspect as a target by using two LSTM surrounding the aspect term.

**TC-LSTM**: Target-connection LSTM upgrades TD-LSTM to capture the connection between aspect and each context word.

**ATAE-LSTM**: Attention-based LSTM with Aspect Embedding joins (i) aspect attention vector with (ii) LSTM hidden vector for the sentiment polarity.

**AF-LSTM**: Aspect-fusion LSTM learns associative relationships between aspect and context words by word-aspect fusion attention.

**HP-LTM**: The authors introduced a hierarchical bidirectional strategy for extracting features.

**MemNet**: Deep memory networks captures the sentimental information without using a recurrent network. This model achieved significant improvements, but with more memory layers.

**Skip-thoughts**: It predicts the surrounding sentences, but its training is time-consuming.

**Byte mLSTM**: This model is one of the most remarkable in recent SC works, since it achieved superior results on customer reviews. It is able to predict the next character (byte) from preceding characters by using mLSTM [127] byte embeddings.

**VIRTUAL ADV:** It is able to capture word-level information from unlabeled data and to learn the parameters. It achieves good performance for semi-supervised and supervised benchmark.

**TopicRNN**: It mixes RNN and topic modeling. It uses long-range dependencies to improve the output word probabilities of documents.

**HN-ATT**: It is a hierarchical network attention model for document classification. Additionally, it implements word-level and sentence-level by using two levels of attention mechanisms.

**Structured Att**: It backwardly infers passes for structured attention, and it uses edge marginal in structured models by the matrix-tree algorithm as attention weights.

### 3) RECURSIVE NEURAL NETWORK (RECNN)
RecNN is a generalization of RNN that applies recursively the same set of weights over a directed acyclic, but in a tree structure input (i.e., words and phrases in a hierarchical structure). RecNN models are linguistically motivated, as they explored tree structures (e.g., syntactic structures) and are aimed to learn complex compositional semantic. The tree structures used for RNNs include constituency tree and dependency tree. On the one hand, in a constituency tree, the words are represented as leaf nodes, the phrases are represented as internal nodes and the root node represents the whole sentence. On the other hand, in a dependency tree, each node represents a word that is connected with other nodes with dependency connections. Generally speaking, in RecNN, the vector (parent) representation of each node is calculated from all of its children using a weight matrix. As shown in Figure 5 (center), RecNN recursively generates parent representations in a bottom-up fashion by combining tokens to produce representations (phrases), and eventually the whole sentence.

RecNN is used for different purposes. It has recently received attention for its compositionality in semantic vector spaces [42]; for instance, in [42], [43] the phrases and sentences are formed in a binary tree. Additionally, it has been used in successful prediction models in cutting-edge domains such as representing phrases. For instance, Qian *et al.* [111] encoded syntactic knowledge in the composition function of RNN. Dong, Wei, Tan, *et al.* [52] transferred a dependency tree of a sentence into a target-specific recursive structure, and get higher level representation based on that structure. Researchers have mainly studied SC on RecNN, proposing models as RAE [43], MVRNN [42], RNTN [36], DRNN [109] and CCAE [128]:

**RAE**: Recursive Autoencoder automatically learns compositionality from sentences. This model could capture the meanings of texts by using predicting structures.

**MVRNN**: Matrix-vector RecNN represents the words as vectors to capture the meaning of long phrases.

**RNTN**: Recursive Neural Tensor Network are used for semantic compositionality of phrases.

**DRNN**: Deep RecNN are able to model sentences by using stacks of multiple recursive layers on top of each other.

**CCAE**: Combinatorial Category Autoencoders capture the compositional aspect of sentences by using combinatorial category grammar operators. It is used for multiple tasks without the need to re-train the main model.

### 4) HYBRID NEURAL NETWORK

The hybrid neural network uses more than one model in one task [125] and, usually, it achieves better accuracy in SC. Thus, several models have been used such as GrConv [129], LSTM-GRNN [25], Conv-GRNN [25] and Tree-LSTM [41]:

**GrConv**: Gated recursive CNN Joins a binary tree (using GRU) with CNN to automatically learn grammatical properties from a sentence.

**Conv-GRNN** and **LSTM-GRNN**: They share a CNN or LSTM with a GRNN to combine the sentence vectors to improve classification on documents.

**Tree-LSTM**: Tree-LSTM learns representations using parsers of LSTM. However, it consumes a considerable training time because it needs to apply predefined syntactic structures.

## IV. EXPERIMENTAL SETUP

This section reports the results of the previously mentioned DL experiments. The first two subsections (A and B) describe (i) the popular datasets used to train/test the models and (ii) the popular word embedding versions, respectively. The metrics used to evaluate the classification performance are described in Subsection C.

### A. DATASETS

Table 3 summarizes the most commonly used datasets of reviews to evaluate SC approaches. It shows (i) the quantitative information (e.g., number of classes, size of the dataset, vocabulary size and length of reviews) and (ii) the qualitative information (e.g., level type, data source and review domain). As shown in the table, the datasets have different classes, sizes, domains, data extracted, labeled/unlabeled reviews, balanced/imbalanced reviews, rating variations of imbalanced, sentence's and token's lengths. These datasets are described below:

**Customer Reviews (CR)**: It is a widely used benchmark dataset[1] (extracted from Amazon) that includes 3,775 full-length customer reviews, where the sentences are labeled as positive or negative [51].

**RT-2k** and **RT-s**: These two datasets[2] consist of movie reviews obtained from Rotten-Tomatoes (RT), where each customer review is classified as positive or negative. The main difference between both of them is in their size. On the one hand, RT-s polarity dataset [10] contains 5,331 positive and 5,331 negative processed reviews of movies. On the other hand, RT-2k contains 1,000 positive snippets and 1,000 negative snippets.

**Stanford Sentiment Treebank (SST)**: SST[3] is an extension of RT-s. It splits into training, development and testing. In SST, the sentences are parsed into parse trees. It includes two benchmark datasets (such as binary version and fine-grained) [36]. The binary version (*SST-2*) is labeled as positive/negative (no neutral reviews), whereas in the fine-grained version (*SST-5*) reviews are classified as very positive, positive, neutral, negative and very negative.

---

These datasets are available at:

[1] http://www.cs.uic.edu/liub/FBS/sentiment-analysis.html
[2] https://www.cs.cornell.edu/people/pabo/movie-review-data/
[3] http://nlp.stanford.edu/sentiment/

TABLE 3. Statistics of the most 21 common freely available review datasets on after tokenization. Class: Number of target classes (opinion polarity). L: Average reviews' length. V: Vocabulary size. Test: Test set size (CV "Cross Validation" means there was no standard train/test split and thus 10-fold CV was used). Dist. (+,-): Lists the class distribution.

| S# | Corpus | | Type | Class | Size | L | Dist.(+,-) | Test | V | Data Source | Review Domain | Format Extracted |
|---|---|---|---|---|---|---|---|---|---|---|---|---|
| 1 | CR | | Sentence-Level | 2 | 3775 | 19 | 0.64/0.36 (Imbalance) | CV | 6k | Amazon | Product | Each of sentence includes a text format and a title with tags of aspect terms |
| 2 | RT-2k | | | 2 | 2000 | 786 | 0.5/0.5 (Balance) | CV | 51k | Rotten-tomatoes | Movie | Each line in each text file corresponds to a single sentence (snippets) |
| 3 | RT-s | | | 2 | 10662 | 20 | 0.5/0.5 (Balance) | CV | 21k | Rotten-tomatoes | Movie | Version of SST-5, with neutral reviews removed and the remaining reviews categorized to either negative or positive. |
| 4 | Stanford Sentiment Treebank | SST-2 | | 2 | 11855 | 18 | (Unlabeled) | 2210 | 18k | Rotten-tomatoes | Movie | For each example in the dataset, there exists only one sentence and a label associated with it. And the labels can be one of (negative, somewhat negative, neutral, somewhat positive, positive). |
| 5 | | SST-5 | | 5 | 9613 | 19 | (Unlabeled) | 1821 | 16k | Rotten-tomatoes | Movie | |
| 6 | IMDb-P | | Document-Level | 2 | 50,000 | 230 | 0.5/0.5 (Balance) | 25k | 392k | IMDb | Movie | This dataset consist of informal reviews. It didn't allow no more than 30 reviews per movie. |
| 7 | IMDb-F | | | 10 | 348,5k | 326 | 07/.04/.05/.08/.11/.15/.17/.12/.18 | 34,85k | 115k | IMDb | Movie | The reviews contain on user ratings (scaled from 0 to 10). |
| 8 | Yelp Challenge | Yelp-13 | | 5 | 335k | 152 | .09/.09/.14/.33/.36 (Imbalance) | 33.5k | 211k | Yelp | Restaurant | each example consists of several review sentences and a rating score range from 1 to 5 (higher is better). |
| 9 | | Yelp-14 | | 5 | 1m | 157 | .10/.09/.15/.30/.36 (Imbalance) | 100k | 476k | Yelp | Restaurant | |
| 10 | | Yelp-15 | | 5 | 1.5m | 152 | .10/.09/.14/.30/.37 (Imbalance) | 150k | 613k | Yelp | Restaurant | |
| 11 | Large Scale | Elec | | 2 | 50,000 | 125 | 0.5/0.5 (Balance) | 25k | 40k | Amazon | Product | The data only includes the text section. |
| 12 | | Yelp-P | | 2 | 598k | 153 | 0.5/0.5 (Balance) | 38k | 116k | Yelp | Restaurant | Each sample is a piece of review text with a binary label (negative or positive). |
| 13 | | Yelp-F | | 5 | 700k | 155 | 0.2/0.2/0.2/0.2/0.2 (Balance) | 50k | 125k | Yelp | Restaurant | |
| 14 | | Amazon-P | | 2 | 4m | 93 | 0.5/0.5 (Balance) | 400k | 395k | Amazon | product | One review has a review title, a review content and a sentiment label. |
| 15 | | Amazon-F | | 5 | 3.65m | 91 | 0.2/0.2/0.2/0.2/0.2 (Balance) | 650k | 357k | Amazon | product | |
| 16 | SemEval | SE14-Lap | Aspect-Level | 3 | 3845 | -- | 0.61/0.18/0.21 (Imbalance) | 800 | -- | Amazon | product | XML tag, in which two attributes ("from and "to") that indicate its start and end offset in the text. |
| 17 | | SE14-Res | | 3 | 3841 | -- | 0.45/0.21/0.34 (Imbalance) | 800 | -- | Yelp | Restaurant | |
| 18 | | SE15-Lap | | 3 | 2500 | -- | 0.57/0.07/0.36 (Imbalance) | 761 | -- | Amazon | Product | XML tag of {Entity # Attribute, polarity}. |
| 19 | | SE15-Res | | 3 | 200 | -- | 0.62/0.05/0.33 (Imbalance) | 685 | -- | Yelp | Restaurant | XML tag of {Entity # Attribute, Opinion-Target-Expression, polarity} |
| 20 | | SE16-Lap | | 3 | 3308 | -- | 0.56/0.07/0.37 (Imbalance) | 808 | -- | Amazon | Product | XML tag of {Entity # Attribute, polarity}. |
| 21 | | SE16-Res | | 3 | 2676 | -- | 0.66/0.04/0.30 (Imbalance) | 676 | -- | Yelp | Restaurant | XML tag of {Entity # Attribute, Opinion-Target-Expression, polarity}. |

**IMDb-P** and **IMDb-F**: These datasets are large scale reviews of movies (full-length). Firstly, the Stanford polarity movie review dataset (IMDb-P)[4] [64] is equally divided between the training and the test reviews (one for positive and another one for negative). Secondly, *IMDb-Full* (*IMDb-F*) is split into ten sentiment labels as presented in [25]. The test set consists of the 10 percent of the whole size.

**Yelp Challenge**[5]: Yelp dataset challenge [25] is obtained from Yelp.com and it contains restaurant reviews, which are organized or classified into five levels of rating. From this dataset, three subsets are constructed: *Yelp-13*, *Yelp-14* and *Yelp-15*. These subsets contain 335,018, 1,125,457 and 1,569,264 samples respectively. Each one of these three is split into (train/dev/test) sets with sizes (80/10/10), respectively.

**Elec**[6]: This dataset is a subset of the large Amazon dataset [130] that consists only of reviews of electronic products [75].

**Large-Scale Datasets:** These datasets[7] were obtained from Yelp.com and Amazon.com. *Yelp-P/F:* Two classification tasks (Full/Polarity) are constructed from this dataset. The first one predicts the number of stars the user has given, and the second one predicts a polarity label by considering stars 1 and 2 as negative, and 3 and 4 as positive. The full dataset has 130,000 training samples and 10,000 testing samples in each star, and the polarity dataset has 280,000 training samples and 19,000 test samples in each polarity, which are also obtained from [24]. *Amazon-P/F:* This product benchmark [24] has two versions (Full/Polarity) as in Yelp review dataset. The authors used the Stanford Network Analysis Project (SNAP) for obtaining on the reviews. The dataset contains 3,650,000 documents for a full dataset and a 4 million documents for a polarity dataset.

**Semantic Evaluation (SemEval)**: This dataset is created for the Aspect-Based Sentiment Analysis (ABSA) task with three polarities (positive, neutral or negative). It is a collection of review datasets (SE-14[8] [131], SE-15[9] [132] and SE-16[10] [133]). These contain several domains and languages. The paper only focus on restaurant reviews and laptop reviews in English. For ABSA task, SE-14 is the most usually used because SE-15 and SE-16 datasets have conflict sentiment for each aspect based on its categories.

B. **WORD EMBEDDING VERSIONS**
Word embeddings are distributed representations of text that encode semantic and syntactic properties of words. They are usually represented as dense and low-dimensional vectors, by applying the previously mentioned approaches. In this section, we briefly discuss the available seven-word embedding datasets that are summarized in Table 4:

**SENNA Embedding:** It is a semantic/syntactic extraction that uses a NN architecture [134]. SENNA behaves very fast and robust (it does not need parsed text) and it is able to label large and noisy corpora. It performs composition over the learned word vectors for classification. In [135], SENNA achieved good results compared with other embedding approaches such as Word2Vec and GloVe.

**Turian Embedding:** This uses word embedding in semi-supervised learning. This embedding covers 268,810 words, each one represented by 25, 50 or 100 dimensions. The weakness of this cannot fully exploit the potential of word vectors.

**HLBL Embedding:** It is a hierarchical log-bilinear model presented by [136]. This embedding covers 246,122 words (one year of Reuters English newswire from August 1996 to August 1997).

**Word2Vec Embedding:** It is an unsupervised learning tool that is provided with two architectures for computing vector representations of words: the continuous bag-of-words and skip-gram. The first one predicts the target word from its context words, while the second does the inverse. Consequently, the skip-gram is better for larger dataset. The phrases were obtained using a simple data-driven approach described in [61]. It is characterized by its speed even with a huge dataset. However, it does not take into account the linguistics of words.

**FastText Embedding:** It is a skip-gram with sub-word character n-grams [137]. It is similar to word2vec and faster for training and evaluation.

**Huang Embedding:** It requires context to disambiguate words. Therefore, Huang incorporates global context to deal with challenges raised by words with multiple meanings. Its size is 100,232 word embedding. The training corpus is: April 2010 snapshot of Wikipedia.

**GloVe Embedding:** it is a global vector that has trained on the nonzero entries of a global word-word co-occurrence matrix [138].

---

These datasets are available at:

[4] http://ai.stanford.edu/amaas/data/sentiment
[5] https://www.yelp.com/dataset/challenge
[6] http://riejohnson.com/cnn_data.html
[7] https://github.com/zhangxiangxiao/Crepe
[8] http://alt.qcri.org/semeval2014/task4/
[9] http://alt.qcri.org/semeval2015/task12/
[10] http://alt.qcri.org/semeval2016/task5/

**TABLE 4.** Description of public released pre-trained word embedding datasets.

| Learning Dense Embeddings | Embedding | Source | Training Corpus | Vocabulary Size | Tokens | Embedding's dimension | Web Resource (URL) | Training Time (epochs) |
|---|---|---|---|---|---|---|---|---|
| Neural Network | SENNA | [134] | Wikipedia | 130,000 | -- | 50 | http://ronan.collobert.com/senna/ | 2 months (50) |
| | Turian | [139] | RCV1 | 268,810 | 1.8B | 25, 50 or 100 | -- | few weeks (50) |
| | HLBL | [136] | Reuters | 246,122 | 37M | 50 or 100 | http://metaoptimize.com/projects/wordreprs/ | 7 days (100) |
| | Word2Vec | [47] | Google News | 3,000,000 | 1B | 300 | https://code.google.com/archive/p/word2vec/ | -- |
| | | [66] | Amazon | 1,000,000 | 4.7B | 50 or 300 | http://sentic.net/ AmazonWE.zip | -- |
| | FastText | [137] | Facebook | -- | -- | -- | https://github.com/facebookresearch/fastText | -- |
| Matrix Factorization | Huang | [140] | Wikipedia | 100,232 | 1.8B | 50 | http://ai.stanford.edu/ehhuang/ | Weeks (50) |
| | GloVe | [138] | Wikipedia | 400,000 | -- | 50, 100, 200 or 300 | http://nlp.stanford.edu/projects/glove/ | -- |
| | | [138] | Twitter | 1,200,000 | -- | 25, 50, 100 or 200 | -- | -- |

## C. PERFORMANCE APPRAISAL

The accuracy (Acc) metric (calculated using Equation 1) [141] is commonly used to measure the performance of SC approaches ([13], [27], [142], [142], [143]). It refers to the proportion of correctly classified samples over the whole samples. Acc is calculated through a confusion matrix. For instance, Table 5 shows the confusion matrix for two classes (Positive, Negative). TP (True Positive) means a positive observation which is predicted as positive, FN (False Negative) means a positive one which is predicted as negative, TN (True Negative) means a negative one which is predicted as negative, and finally FP (False Positive) means the observation is negative and is predicted as positive.

**TABLE 5.** The confusion matrix.

|  |  | Predicted | |
|---|---|---|---|
|  |  | Positive | Negative |
| Actual | Positive | TP | FN |
| | Negative | FP | TN |

$$Acc = \frac{TP+TN}{TP+TN+FP+FN} \quad (1)$$

Although Acc has been chosen by most of the researchers to measure the performance of DL-based SC approaches, other metrics such as precision, recall and f-measure could provide better insights. These other metrics are not used in DL-based SC approaches, although some researchers have used them in other SA tasks such as aspect extraction ([27], [142]). Additionally, other interesting metrics, such as weighted kappa [144], are used to distinguish between disagreements (misclassification) of negative to positive. A study by Novielli et al [145] analyzed misclassification between sentiment tools in software engineering and concluded that even though there are disagreements between the tools, it is minimal and dataset dependent. However, weighted kappa is not used in DL-based sentiment classification approaches. Consequently, and since our comparison is a literature-based one, we were forced to only use the Acc metric to compare the different DL-based SC approaches. Researchers in this field could be motivated to use and address other metrics in their experiments in order to have the possibility to apply other metrics for broader comparisons.

## V. A COMPARISON OF DL-BASED SC APPROACHES

In this section, we compare the performance of SC techniques in three domains: product, movie and restaurant reviews. As it was previously mentioned, these three domains entail interesting challenges and constitute the most common domains in review mining. The performance of the state-of-the-art approaches on the available datasets are summarized in Tables 6-8. The provided results in these tables are reproduced under the same measure (accuracy) and conditions that they were state in the original papers.

***Product Reviews:*** The analysis of product reviews is very important for manufacturers/companies to understand the customer opinions as well as for the customers to know other consumers' feedbacks. Unfortunately, the extracted reviews from this domain contain a considerable amount of noisy data. So, a huge effort in preprocessing the datasets is needed. The results of applying SC approaches on the seven datasets (CR, Amazon-P, Amazon-F, Elec, SE14-Lap, SE15-Lap and SE16-Lap) are summarized in Table 6. For each data set, the accuracy resulted from each approach is added to the table. As shown in table 6, Byte mLSTM [32] model significantly outperformed other methods on CR dataset, while TreeNet achieved better performance and generalization with fewer parameters. In Byte mLSTM, a byte-level language model trained on the large product review dataset is used to obtain sentence representations. For the Elec datasets, ADV-LM-LSTM+IMN [146] gave the best result, where ADV-LM-LSTM [68] provides a performance competitive with the current best result for the configuration of supervised learning. In [146], they added the IMN to improve the performance. VAT-LM-LSTM [91] has lower performance than VIRTUAL ADV [68] on Elec, while VAT-LM-LSTM is the most reliable result for the comparison as expected by [146]. In [68], they applied perturbations to the initial pre-trained word embeddings in conventional LSTM. On Amazon (polarity/full), DPCNN [155] achieved the highest accuracy, although it gave lower performance in other domains (e.g., the restaurant domain shown in Table 8). The DPCNN model develops CNN for capturing the n-gram features (i.e., can capture n-gram information of different sizes without manually setting the convolution kernel size). On SE14-Lap, BBLSTM-SL model [154] achieved the second best reported result, only after the significantly LCR-Rot [149] model, where this model is able to better represent the sentiment aspect, especially when the aspect contains multiple words. On SemEval 15/16 datasets of laptop domain, the system that is presented by [156] is competitive. This system is scalable and able to process a high volume of opinion-based documents in real-time.

TABLE 6. Comparison with state-of-the-art results from the literature on the most widespread available *Product* review datasets on SC.

| DS | Model | Acc | DS | Model | Acc | DS | Model | Acc | DS | Model | Acc |
|---|---|---|---|---|---|---|---|---|---|---|---|
| CR | DisSent [96] | 84.9 | Amazon-P | Char-CRNN [55] | 94.10 | SE14-Lap | PF-CNN [83] | 70.06 | SE16-Lap | Senti [133] | 74.28 |
| CR | 3W-CNN [77] | 85.8 | Amazon-P | FastText [137] | 94.60 | SE14-Lap | GRNN-G3 [100] | 71.47 | SE16-Lap | [147] | 74.58 |
| CR | CNN-multichannel [40] | 85.0 | Amazon-P | BiLSTM + KNN [148] | 94.70 | SE14-Lap | Feature-enhanced SVM [149] | 72.10 | SE16-Lap | LeeHu [133] | 75.91 |
| CR | CNN-Rule-q [120] | 85.3 | Amazon-P | Char-level CNN [24] | 95.07 | SE14-Lap | IAN [101] | 72.10 | SE16-Lap | AUEB [133] | 76.9 |
| CR | QuickThoughts [98] | 86.0 | Amazon-P | BiLSTM + KNN [148] | 95.30 | SE14-Lap | MemNet [35] | 72.37 | SE16-Lap | NileT [133] | 77.40 |
| CR | AdaSent [30] | 86.3 | Amazon-P | Region-embedding [150] | 95.30 | SE14-Lap | Coattention-MemNet [143] | 72.90 | SE16-Lap | IHS-R [133] | 77.90 |
| CR | SuBiLSTM [151] | 86.5 | Amazon-P | SANet [84] | 95.48 | SE14-Lap | DMN [102] | 72.95 | SE16-Lap | ECNU [152] | 78.15 |
| CR | MultiTask [97] | 87.7 | Amazon-P | VDCNN [153] | 95.69 | SE14-Lap | Coattention-LSTM [143] | 73.50 | SE16-Lap | IIT-T [133] | 78.40 |
| CR | TreeNet [113] | 88.4 | Amazon-P | word-CNN [114] | 96.21 | SE14-Lap | BBLSTM-SL [154] | 74.90 | SE16-Lap | INSIG [133] | 78.40 |
| CR | Byte mLSTM [32] | 91.4 | Amazon-P | DPCNN [155] | 96.68 | SE14-Lap | LCR-Rot [149] | 75.24 | SE16-Lap | [156] | 81.08 |
| Elec | oh-2LSTMp [157] | 93.92 | Amazon-F | Char-CRNN [55] | 59.20 | SE15-Lap | wnlp [132] | 72.07 | | | |
| Elec | One-hot CNN [157] | 94.13 | Amazon-F | Char-level CNN [24] | 59.57 | SE15-Lap | EliXa [132] | 72.91 | | | |
| Elec | iAdvT-Text [91] | 94.42 | Amazon-F | FastText [137] | 60.20 | SE15-Lap | TJUdeM [132] | 73.23 | | | |
| Elec | One-hot bi-LSTM [157] | 94.45 | Amazon-F | BiLSTM [148] | 60.30 | SE15-Lap | LT3 [132] | 73.76 | | | |
| Elec | VAT-LM-LSTM [68] | 94.46 | Amazon-F | Region-embedding [150] | 60.80 | SE15-Lap | [50] | 75.87 | | | |
| Elec | LM-LSTM+IMN [146] | 94.52 | Amazon-F | SANet [84] | 61.33 | SE15-Lap | [147] | 76.54 | | | |
| Elec | VIRTUAL ADV [68] | 94.60 | Amazon-F | VDCNN [153] | 63.00 | SE15-Lap | Lsislif [132] | 77.87 | | | |
| Elec | L$_{ML}$ [90] | 94.76 | Amazon-F | DCCNN-ATT [5] | 63.00 | SE15-Lap | ECNU [152] | 78.29 | | | |
| Elec | iVAT-LSTM [91] | 94.82 | Amazon-F | word-CNN [114] | 63.76 | SE15-Lap | sentiue [158] | 79.34 | | | |
| Elec | ADV-LM-LSTM+IMN [146] | 94.86 | Amazon-F | DPCNN [155] | 65.19 | SE15-Lap | [156] | 85.89 | | | |

*Movie Reviews*: Movie reviews constitute the most challenging domain within review mining. Data is usually extracted from Rotten-tomatoes and the Internet Movie Database (IMDb). The extracted reviews do not offer a clear idea of what are factual information and which are mainly opinions. Thus, movie reviews are apparently harder to classify than product reviews, since product reviews have less specific features [18], [54]. This is why the movie domain is experimentally convenient and has larger online reviews [17]. Table 7 gathers a list of approaches and their results on six public movie review datasets. As shown in this table, DAN [108] model has achieved the worst performance on most of the datasets because it did not initialize with pre-trained embedding or randomly. The Byte mLSTM model is competitive with more complex models on the SST-2 and it outperforms the state-of-the-art models. But with SST-5 and IMDb-P datasets, it did not perform well. This model trained a simple logistic regression classifier with L1 regularization. It achieved the second best result on the RT-2k dataset after AC-BiLSTM model. On RT-2K, AC-BiLSTM [159] achieved the highest accuracy on LSTM based models and entails an improvement of 10.8, 9.7, 9.2 and 6.1 over TE-LSTM, SA-LSTM, SATA TLSTM, and Byte mLSTM, respectively. AC-BiLSTM model is based on a bidirectional LSTM and attention mechanism. Bidirectional LSTM is adopted to access the preceding and succeeding context representations. While the attention mechanism is intended to give focus on the information produced by the hidden layers of BiLSTM. CRAN [115] gives best results on RT-S dataset, where it combines RNN with the convolutional attention model to resolve aspect-level sentiment analysis tasks as well. The BCN ELMo [160] method achieved better results than other methods on SST-5. These methods combined the Biattentive Classification Network (BCN) with Embeddings from

**TABLE 7.** Comparison with state-of-the-art results from the literature on the most widespread available *Movie* review datasets on SC.

| DS | Model | Acc | DS | Model | Acc | DS | Model | Acc |
|---|---|---|---|---|---|---|---|---|
| RT-2k | DAN [108] | 80.3 | SST-2 | MVCNN [118] | 89.4 | IMDb-P | RCRN [164] | 92.8 |
| RT-2k | CNN-Rule-q [120] | 81.7 | SST-2 | TE-LSTM [94] | 89.6 | IMDb-P | Byte mLSTM [32] | 92.88 |
| RT-2k | TE-LSTM [94] | 82.2 | SST-2 | NSE [165] | 89.7 | IMDb-P | CNN+tvEmb [166] | 93.49 |
| RT-2k | QuickThoughts [98] | 82.4 | SST-2 | IRAM [116] | 90.1 | IMDb-P | Ensemble [167] | 93.51 |
| RT-2k | MultiTask [97] | 82.5 | SST-2 | CT-LSTM [168] | 90.2 | IMDb-P | TopicRNN [33] | 93.72 |
| RT-2k | SA-LSTM [169] | 83.3 | SST-2 | BCN [164] | 90.3 | IMDb-P | ADV-LM-LSTM+IMN [146] | 93.93 |
| RT-2k | VIRTUAL ADV [68] | 83.4 | SST-2 | AR-Tree [164] | 90.4 | IMDb-P | oh-2LSTMp [157] | 94.06 |
| RT-2k | SATA TLSTM [170] | 83.8 | SST-2 | RCRN [171] | 90.6 | IMDb-P | VIRTUAL ADV [68] | 94.09 |
| RT-2k | Byte mLSTM [32] | 86.9 | SST-2 | SATA TLSTM [170] | 91.3 | IMDb-P | iVAT-LSTM [91] | 94.34 |
| RT-2k | AC-BiLSTM [159] | 93.0 | SST-2 | Byte mLSTM [32] | 91.8 | IMDb-P | L$_{ML}$ [90] | 95.68 |
| RT-s | CNN [40] | 81.5 | SST-5 | TE-LSTM [94] | 52.6 | IMDb-F | SVM + TextFeatures [25] | 40.5 |
| RT-s | RR-CNN [172] | 81.6 | SST-5 | Byte mLSTM [32] | 52.9 | IMDb-F | CNN [40] | 40.6 |
| RT-s | SA-SNN [44] | 82.1 | SST-5 | NTI [173] | 53.1 | IMDb-F | SVM + Bigrams [25] | 40.9 |
| RT-s | HS-LSTM [174] | 82.1 | SST-5 | CT-LSTM [168] | 53.6 | IMDb-F | Conv-GRNN [25] | 42.5 |
| RT-s | SFCNN [175] | 82.7 | SST-5 | AR-Tree [164] | 53.7 | IMDb-F | FastText [137] | 45.2 |
| RT-s | AdaSent [30] | 83.1 | SST-5 | IRAM [116] | 53.7 | IMDb-F | LSTM-GRNN [25] | 45.3 |
| RT-s | AC-BiLSTM [159] | 83.2 | SST-5 | Gumbel Tree-LSTM [117] | 53.7 | IMDb-F | Structured Att [126] | 49.2 |
| RT-s | ACNN [176] | 83.4 | SST-5 | RCRN [164] | 54.3 | IMDb-F | HN-ATT [87] | 49.4 |
| RT-s | TreeNet [113] | 83.6 | SST-5 | SATA TLSTM [170] | 54.4 | IMDb-F | H-CRAN [115] | 50.2 |
| RT-s | CRAN [115] | 83.8 | SST-5 | BCN ELMo [160] | 54.7 | IMDb-F | TCV [161] | 50.5 |

Language Models (ELMo) word representations to encode sentences to pass through classifiers. TCV [161] in movie reviews outperforms the other models, unlike restaurant reviews. This model proposed Text Concept Vector (TCV) for the text representation which extracts the concept level information of text. At the document-level IMDb-P task, the LML [90] model is competitive with more complex models. This model applied entropy minimization to unlabeled data in an unsupervised way. Thus, it outperforms VIRTUAL ADV and iVAT-LSTM. H-CRAN [115] achieved the second best result on the IMDb-F dataset after TCV model. TCV model obtains a substantial improvement in classification accuracy compared with the more complex methods, for the reason given above.

*Restaurant Reviews:* Table 8 gives the results of different SC models on different restaurant datasets and it can be perceived that H-CRAN model achieved good performance on the three yelp datasets (Yelp13, Yelp-14 and Yelp-15). For these three datasets, H-CRAN [115] outperforms LSTM-GRNN [25] model by 3.6%, 4.4% and 5.4%, respectively. H-CRAN used a similar hierarchical architecture than LSTM-GRNN, but with an additional attention mechanism to extract salient words in sentences and salient sentences in the document. ULMFiT [162] model outperforms other models on the large scale datasets Yelp-P and Yelp-F. ULMFiT model is well known for their transfer learning capabilities and good generalizing for various NLP tasks across different domains.

On SE14-Res, DMN [102] implemented multiple attention mechanism on LSTM and it outperforms all others. FastText [137] model is widespread on document-level SC of this domain. FastText has been shown to train quickly and to achieve prediction performance comparable to RNN embedding model for SC. Compared to FastText, many models gave superior performance across the board. ECNU [152] model achieved the second best reported result on the SE15-Res dataset, only after the significantly slower Sentiue [158] model. It is also competitive on SemEval 15/16 datasets. ECNU used a lexicon-based approach based on domain-specific lexicons, while Sentiue used a MaxEnt classifier with a set of features including lexicons. On SE16-Res, XRCE [163] model outperforms all other models. This model used symbolic parser designed with special lexicon and combined with SVM.

Furthermore, we have compared the addressed DL-based SC approaches on the three domains products, movies and restaurants reviews; by using the three classes of factors suggested in Section III. The average of the approaches performances is used on the three domains for the different factors. These factors-based comparisons are discussed in the following subsections.

TABLE 8. Comparison with state-of-the-art results from the literature on the most widespread available _Restaurant_ review datasets on SC.

| DS | Model | Acc | DS | Model | Acc | DS | Model | Acc | DS | Model | Acc |
|---|---|---|---|---|---|---|---|---|---|---|---|
| Yelp-13 | CNN [40] | 62.7 | Yelp-15 | Paragraph Vector [47] | 60.5 | Yelp-F | SWEM [177] | 63.79 | SE15-Res | EliXa [132] | 70.05 |
| | Conv-GRNN [25] | 63.7 | | SVM + TextFeatures [25] | 62.4 | | FastText [137] | 63.90 | | UMDuluth [132] | 71.12 |
| | FastText [137] | 64.2 | | SVM + Bigrams [25] | 62.4 | | SANet [84] | 63.97 | | SIEL [132] | 71.24 |
| | LSTM-GRNN [25] | 65.1 | | CNN [40] | 64.5 | | VDCNN [153] | 64.72 | | wnlp [132] | 71.36 |
| | TCV [161] | 67.8 | | Conv-GRNN [25] | 66.0 | | Region-embedding [150] | 64.90 | | UFRGS [132] | 71.71 |
| | Structured Att (doc) [126] | 67.8 | | FastText [137] | 66.6 | | [178] | 65.83 | | LT3 [132] | 75.02 |
| | Structured Att [126] | 68.0 | | LSTM-GRNN [25] | 67.6 | | DCCNN-ATT [5] | 66.00 | | lsislif [132] | 75.50 |
| | HN-ATT [87] | 68.2 | | HN-ATT [87] | 71.0 | | word-CNN [114] | 67.61 | | [156] | 77.94 |
| | Structured Att (both) [126] | 68.6 | | TCV [161] | 71.5 | | DPCNN [155] | 69.42 | | ECNU [152] | 78.10 |
| | H-CRAN [115] | 68.7 | | H-CRAN [115] | 73.0 | | ULMFiT [162] | 70.02 | | sentiue [158] | 78.69 |
| Yelp-14 | Paragraph Vector [47] | 59.2 | Yelp-P | FastText [137] | 95.70 | SE14-Res | Coattention-LSTM [143] | 78.80 | SE16-Res | [147] | 81.62 |
| | CNN [40] | 61.4 | | VDCNN [153] | 95.72 | | PG-CNN [83] | 78.93 | | INSIG [133] | 82.07 |
| | SVM + Bigrams [25] | 61.6 | | SWEM [177] | 95.81 | | PF-CNN [83] | 79.20 | | [50] | 83.18 |
| | SVM + TextFeatures [25] | 61.8 | | DCNN [179] | 96.04 | | GRNN-G3 [100] | 79.55 | | AUEB [133] | 83.24 |
| | Conv-GRNN [25] | 65.5 | | Region-embedding [150] | 96.40 | | Coattention-MemNet [143] | 79.70 | | ECNU [152] | 83.59 |
| | FastText [137] | 66.2 | | DCCNN-ATT [5] | 96.50 | | Feature-enhanced SVM [149] | 80.89 | | IHS-R [133] | 83.94 |
| | LSTM-GRNN [25] | 67.1 | | [178] | 96.58 | | MemNet [35] | 80.95 | | NileT [133] | 85.45 |
| | TCV [161] | 69.2 | | word-CNN [114] | 97.10 | | BBLSTM-SL [154] | 81.30 | | IIT-T [133] | 86.73 |
| | HN-ATT [87] | 70.5 | | DPCNN [155] | 97.36 | | LCR-Rot [149] | 81.34 | | [156] | 87.10 |
| | H-CRAN [115] | 71.5 | | ULMFiT [162] | 97.84 | | DMN [102] | 81.43 | | XRCE [163] | 88.13 |

## A. DATA PREPARATION BASED COMPARISON

The comparison of the factors in data preparation phase on the three review domains shows the average performance of SC approaches for each factor. As shown in Figure 6, the factors magnitudes, annotation, granularity and equilibrium are able to change performance of SC as follows. First, the results show that the more number of classes in a review-dataset, the lower performance of the applied models, as shown in Figure 6(a). This comparison is based on the Large-Scale Dataset from 2 and 5 classes, while in 3-classes, the comparison is based on the SemEval datasets. Also, a slight superiority of restaurant-reviews over product-reviews is noted. The movie-reviews datasets are not used as no ternary classes are available for this domain. Second, in Figure 6(b), using Rottentomatoes movie reviews to compare both labeled (e.g., RT-2 and RT-S) and unlabeled data (e.g., SST-2 and SST-5), the comparison shows that labeled-data dominate unlabeled-data on 2 and 5 classes. Third, these domains are compared on document granularity level. Document-level achieves the highest performance from others in the same class (e.g., binary class) as in Yelp-p, Amazon-p, Elec, IMDb-p, SST-2 and CR. In Figure 6(c), three datasets from the binary class are compared at document-level. In this level, restaurant-reviews beat the others, which is less noise data. Fourth, movie-review datasets (e.g., IMDb-F and IMDb-P) have high performance on balanced-data. This may happen due to the big difference among number of classes they have. So, product reviews are compared at the same class but not the same level (e.g., CR and Elec). Also, balanced-data is outdone. On restaurant reviews, the same class, level and the same size approximately (e.g., Yelp-14 and Yelp-F) are compared. The results showed that the balanced-data outperformed, as shown in Figure 6(d).

## B. FEATURE REPRESENTATION BASED COMPARISON

This section compares two factors in the feature representation phase: Learning word embedding and training corpus type. At the end of this comparison, we will be able to answer the following two questions: "*What is the most appropriate way to learn dense embedding in SC task?*" and *"What is the best data type for training in SC task?"* For the first factor, we compare *Word2Vec* of neural network and *GloVe* of matrix factorization that are the most prevalent (due to their high performance on SC) on review domains. Figure 7(a) offers a comparison on the 3 domains in document-level on different classes (e.g., IMDb-F, Amazon-P/F and Yelp-13). The figure shows that matrix factorization-based method (i.e., GloVe) is superior to the other. In the second factor, we provide a comparison on Word2Vec for the two corpora Amazon and Google in movie-reviews and product-reviews domains (as Word2Vec is more prevalent in both domains). As shown in Figure 7(b), superiority of Amazon corpus-based methods is found in the two review domains (products and movies) over Google corpus-based methods. This happens due to the big opinionated information in Amazon corpus.

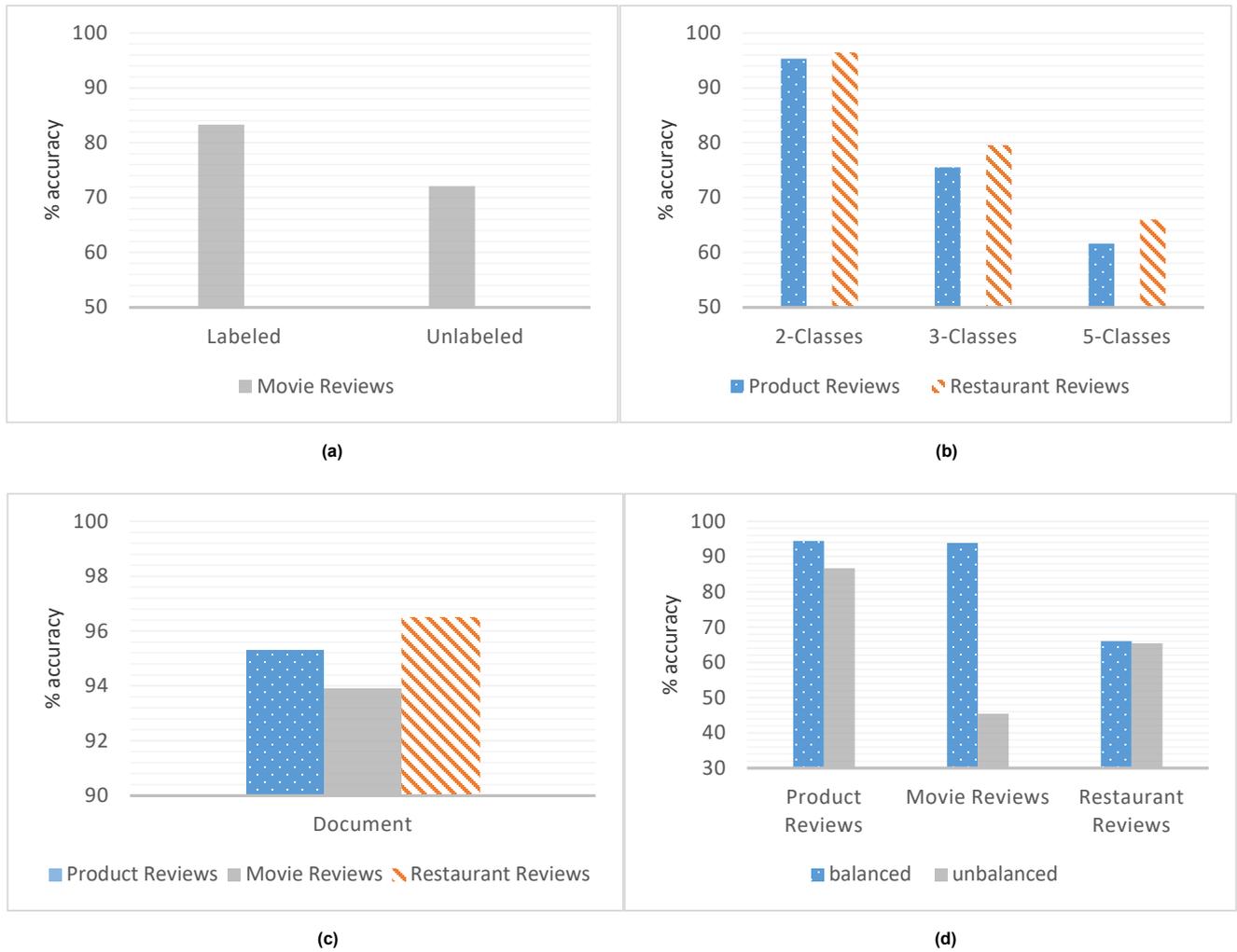

FIGURE 6. Comparison of SC approaches on the review domains based on the *data preparation* based factors.

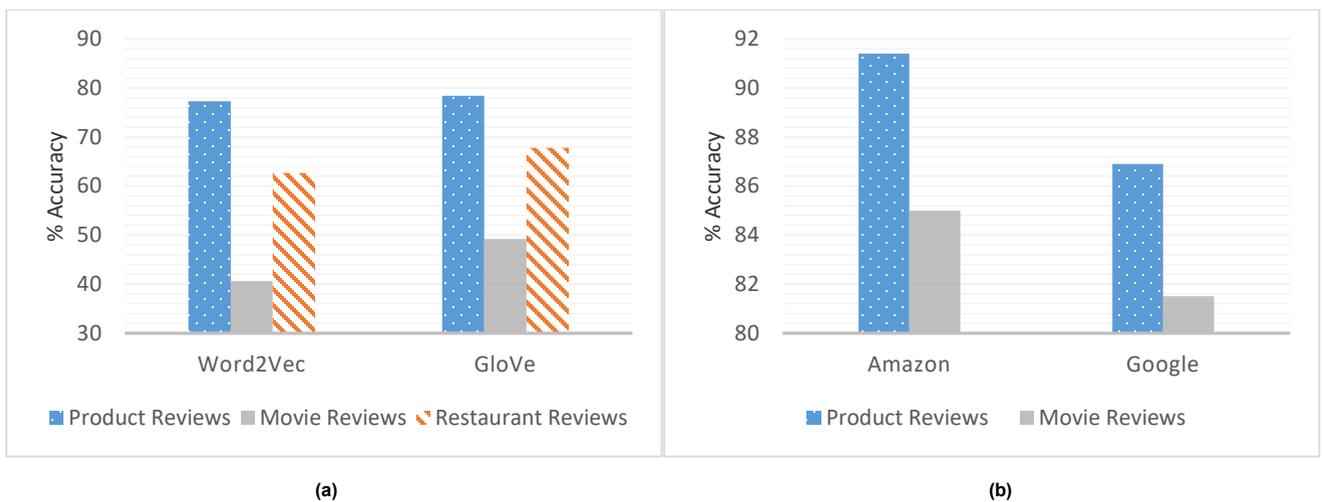

FIGURE 7. Comparison of SC approached on the review domains based on *word embedding*.

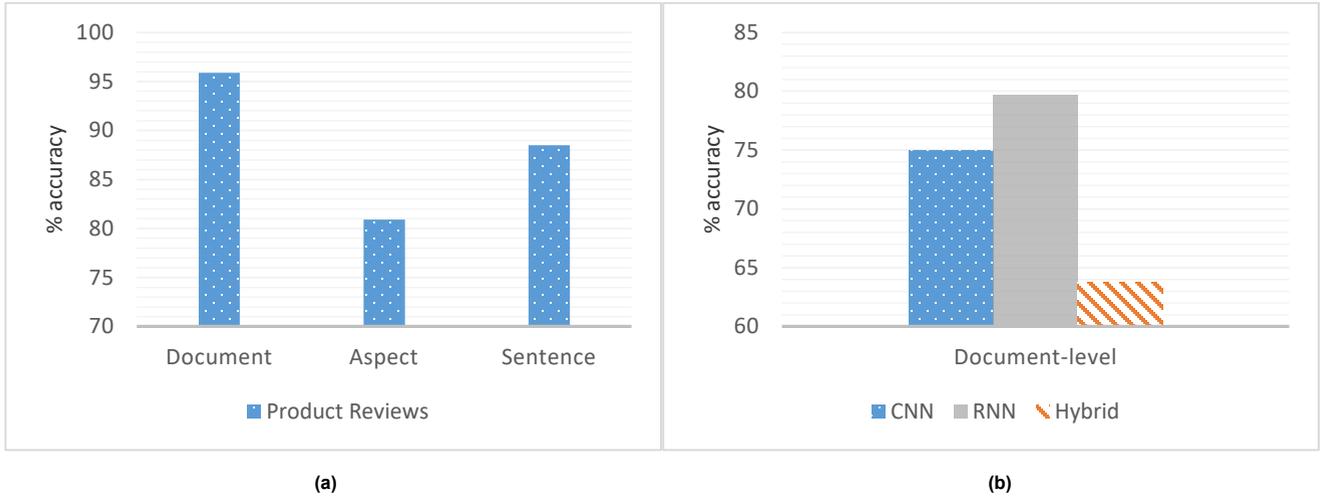

**FIGURE 8.** Comparison of SC approached on the review domains based on the used *DL techniques*.

### C. TECHNIQUE BASED COMPARISON

This section compares SC approaches based on the two following factors: determining the level of the granularity and the architecture of the DL model. Figure 8 shows that the two factors affect the performance of SC. In the first factor, we compare the results of the DL-methods from different review domains at the 3 levels: document-level, sentence-level and aspect-level, as shown in Figure 8(a). We did not find a DL method that has been applied to the three levels at the same time. Therefore, we compare the highest average performance of the datasets at each level (i.e., using Yelp-P/Amazon-P of document-level, SST-2/CR of sentence-level and SE16-Res/ SE16-Lap of aspect-level). We have noticed that document-level has the best performance as compared to the others. Therefore, in the second factor, we are concerned with comparing DL architectures (e.g., CNN, RNN, RecNN and Hybrid) on this level, as shown Figure 8(b). We have found that RNN is better than the others. In this comparison, RecNN methods are excluded because these methods are rarely applied (for its lack of quality at this level). In [155], they tried on the level of documents in movie and product reviews such as IMDb-P and Elec respectively, in binary class. We also find a clear advantage of RNN over CNN and hybrid.

### D. ANALYSIS OF THE RESULTS

The results of SC approaches on the three different domains are summarized in Tables 6-8. This section analyzes the most important and recent DL-methods at the three different granularity levels: aspect, sentence and document:

At the **aspect-level** SC, many researchers use RNN (e.g., GRNN-G3 [100], MemNet [35], IAN [101], Coattention-LSTM/MemNet [143] and BBLSTM-SL [154]). These approaches were applied on SemEval-14 (Restaurant and Laptop datasets) for the ABSA task. Recently, Coattention-LSTM/MemNet used location information. These are more effective than MemNet and IAN over Laptop datasets. But, in restaurant datasets, MemNet outperforms both Coattention-LSTM/MemNet and IAN. MemNet model did not depend on sentiment lexicon or syntactic parser, and it is computationally efficient. BBLSTM-SL model is more effective than the previous models, since it utilized sentiment lexicon as a feature for words. Also, CNN has a bit of luck for aspect-based SC task (e.g., PG-CNN [83], PF-CNN [83]). The results of these approaches have the worst performance as compared to RNN approaches.

At **sentence-level** SC, all DL models have been used widely. There are two subtasks on this level (subjectivity classification and sentence-level SC), but we only focused on the second one which is the most common and used with all NN architectures: CNN approaches (CNN-Rule-q [120]), RNN approaches (QuickThoughts [98], MultiTask [97], Byte mLSTM [32]) and RecNN approaches (AdaSent [30], TreeNet [113]). It is interesting to say that CNN-Rule-q method is learned simultaneously during the training from labeled instances (comparing product and movie reviews of the three architectures). However, it has the worst performance as compared to the others. In RecNN, TreeNet is more effective than AdaSent on the two domains. The RNN approaches are also widespread in this level of SC. The Byte mLSTM model is applied in the document-level, but it is more robust in sentence-level (especially binary corpus). It achieves better results than QuickThoughts and MultiTask on the two domains (Products and Movies).

At **document-level** SC, both RNN and CNN are competitors, although RNN outperforms CNN. The comparison on the Product and Restaurant datasets show that DPCNN [155] (a CNN method) consistently outperforms VDCNN [153] and Hybrid approaches such as (DCCNN-ATT [5] and word-CNN [155]). The RNN model is common on the Product and Movie datasets together (e.g., oh-2LSTMp [157], LML [90], iAdvT-Text [91], LM-LSTM+IMN [146]). The most robust in performance from these approaches is LML on Movie reviews while LM-LSTM+IMN [146] on Product reviews. Also, the RNN is publically on the implementations of Movie and Restaurant reviews together such as HN-ATT [87], Structured Attention [126]. HN-ATT [87] achieves better results than the other method on Movie datasets. We have concluded from this analysis that RNN also excels at the aspect level and sentence level over the others.

## VI. DISCUSSION & OPEN ISSUES

This section presents extra factors that have not been previously mentioned and that are expected to affect the performance of SC approaches. Finally, some open issues in SC are introduced.

There are general factors in **data preparation** as domain determination, underlying language and corpus type. In *domain determination*, there are many domains on SA field: blogs, reviews (hotels, electronics, restaurants, etc.), social media and others. In our work, we focused on the review domains since it is the most prevalent in SC. In the selected field, there are also some aspects to take into account, for instance, reviews are not easily processed because they may contain irony or slangs. Moreover, the *underlying language* is an important factor for changing the performance in the SC task. For some languages, like English, there are many tools that enhance the performance of the classification problem. Sometimes, a model gives good results with a language, but not so well with others. For instance, A Multinomial Naïve Bayes model gave the best results for English, Support Vector Machine model for Dutch, and Maximum Entropy model for French as reported in [3]. We have used English because it is a widespread language.

Besides, the factors presented in Table 3 affect the performance of the DL-based SC approaches. That is, corpus's size, data source (Is the data extracted from different websites similar?), CV-based split (Some of the experiments were based on cross validation, others were divided into specific proportions), review length and vocabulary size, and the format of the data extracted (some of them were text and others were xml) have an important influence in the analysis.

**Corpus's size:** As noted in the results, large scale datasets (e.g., Yelp-P and Amazon-P) achieved better results than smaller ones (e.g., CR and RT.2k). Corpus size and corpus domain have an effect on the system performance [16].

**Data source:** We focused on corpora that built on a rating system such as Amazon, Yelp, IMDb and rottentomatoes because they are adequate for automatic labeling and professional review sites.

**Cross-validation-based split:** Some of the experiments were based on CV, others were divided into specific proportions. CV is used to estimate the predictive performance of the models more accurate than dividing data by specific proportions (e.g., 20% test-set and 80% train-set). Therefore, we have noted that the results with cross-validation are relatively less. For instance, SST-2 and RT-s are nearly identical, but SST-2 is more accurate than RT-s with cross-validation.

**Review's length and vocabulary size:** We have noticed that RT.2k achieved higher performance than RT.s, although the size of corpus is less. This is because the length of the reviews and variety of vocabulary in RT.2k are higher than in RT.s.

**Format extraction:** Reviews have different formats. Some reviews are text and others are XML. Consequently, we have noticed that results of the two format are not the same.

Furthermore, there are other three factors that are key on feature representation: training corpus size, accuracy of embedding and method type. For the first factor, the increase of corpus size positively affects the performance of SC. Thus, word embedding versions must be trained with large data ([180], [70]). Even with similar domain, a huge corpus proved its effectiveness [178]. Severyn *et al.* [181] trained word embedding using Word2Vec on 50 million tweets. Word2Vec method is learned on 408 million words of Wikipedia in [180]. For the second factor, some algorithms have been proposed to increase the accuracy of pre-trained word embedding [181, 182], as the amendment increases the accuracy in SC. For the third factor, method type, there are three types of feature representation affect the performance of DL-based methods: supervised, semi-supervised and unsupervised. As for SC, there are several DL models that have been applied to the three classifiers of feature representation, as shown in Table 9. First, supervised methods require more labeled data to work well. Second, semi-supervised methods use large unlabeled data with small labeled data to reduce sparse data. Their approaches improve generalization accuracy, but it consumes time to adapt with supervised systems. Third, unsupervised methods train on unlabeled data separately, but it is not sufficient for SC [129]. It used to improve classification accuracy [91]. We think these factors are effective, so we will discuss them in-depth in a future work.

**TABLE 9.** Distribution of papers based on SC with the three types of feature representations.

| Method Type | Articles' References |
| --- | --- |
| supervised | [129], [125], [30], [36], [41], [109], [43], [182], [183], [128], [42], [151] |
| semi-supervised | [166], [169], [68], [90], [146] |
| *unsupervised* | [155], [184], [47], [38], [108], [75], [32], [185], [186], [24], [137], [98] |

The proposed factors that affect the performance of SC approaches give researchers a whole insight to enhance the performance of their SC approaches. It also highlights three open issues that could be listed as follows.

- Reviews collected from various resources may contain irony, colloquial language, abbreviation, slangs, noisy, wrongly spelt, or not grammatical text. There is no automatic system for solving these gabs. So, data preparation phase consumes maximum time to convert data into a structured format.
- There is a lack of SC tools in non-English languages. For instance, Arabic has no enough tools, besides its parser complexity and synonymous dis-ambiguity make the classification problem a challenge. Thus, large companies suffer from this problem when analyzing customers' opinions of different languages. Furthermore, there is no system that works equally well on various review domains, due to the different orientation of opinion words for each field.
- Few attempts were made to create word embedding that combined the advantages of both neural network and matrix factorization, and increase accuracy of embedding.

## VII. CONCLUSIONS

This paper is a comprehensive survey for the state-of-the-art in Deep Learning (DL) methods based on Sentiment Classification (SC). Our results showed that, although DL models give high performance and outperform others, many factors could affect the performance of these DL-based SC approaches. Besides, we have provided different literature-based comparisons to find the most significant factors in the three phases: data preparation, feature representation and classification techniques. The comparisons are conducted using more than 100 techniques and applied to 21 textual datasets.

In the *data preparation* based comparisons, we have obtained that: (i) The greater number of classes in a dataset, the more challenging SC problem; (ii) Labeled-datasets achieve better results than unlabeled-datasets; (iii) When comparing different review domains (products, movies and restaurants) on document level, it is found the superiority of the restaurant reviews while the movie reviews have the least accuracies; and (iv) Balanced datasets outperformed imbalanced-datasets. In the feature representation based comparisons, we have deduced that: (i) Word embedding based matrix factorization (i.e., GloVe) is better than word embedding based neural network in the performance for SC (i.e., Word2Vec); and (ii) Training data on Amazon corpus with Word2Vec dense embedding is more appropriate for SC than Google corpus. In the classification technique based comparison, we have found that: (i) the three most common NN architectures (RNN, CNN and Hybrid NN) are widely used to solve the three levels (document, sentence and aspect) of SC, and they perform well with the document-level SC. RNN achieved the highest results as compared to the others; and (ii) the performance of DL-based models on the aspect-level SC is still a challenge and needs much more effort to be solved.

In this survey, we only focused on comparing results of DL models on benchmark datasets of customers' reviews. In the future, we hope to compare DL models on different word embedding versions in-depth. We expect that these kinds of in-depth comparisons and analysis provide researchers in this field by more factors that affect the performance of SC.

## ACKNOWLEDGMENT

The authors would like to thank the European Regional Development Fund (ERDF) and the Galician Regional Government, under the agreement for funding the Atlantic Research Center for Information and Communication Technologies (AtlantTIC), and the Spanish Ministry of Economy and Competitiveness, under the National Science Program (TEC2017-84197-C4-2-R).

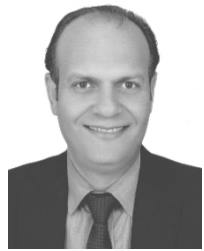

**MOHAMMED KAYED** received the M.Sc. degree in Computer Science from Minia University, Minia, Egypt, in 2002 and the Ph.D. degree in Computer Science from Beni-Suef University, Beni-Suef, Egypt, in 2007.

From 2005 to 2006, he was a Research&Teaching Assistant in Department of Computer Science and Information Engineering at the National Central University, Taiwan. Since 2007, he has been an Assistant Professor with Math&CS Department, Faculty of Science, Beni-Suef University, Beni-Suef, Egypt. He is currently an Associate Professor and Head of Computer Science Department, Faculty of Computer and Artificial Intelligence, Beni-Suef University, Egypt. He is the author of more than 18 articles. His research interests include Web mining, Opinion Mining, Informarion Extraction and Information Retrieval.

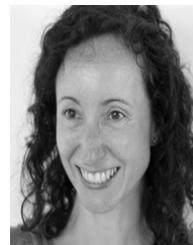

**REBECA P. DÍAZ REDONDO** is an Associate Professor at the Telematics Engineering Department at the University of Vigo. She is currently working on defining appropriate architectures for distributed and collaborative data analysis, especially thought for IoT solutions, where computation must be on the edge of the network (Fog Computing). Rebeca has participated in more than 40 projects and 25 works of technological transfer through contracts with companies and/or public institutions. She is currently involved in the scientific and technical activities of several National and European research & educative projects.

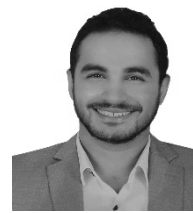

**ALHASSAN MABROUK** is a teaching assistant at Mathematics and Computer Science Department at Beni-Suef University. He received the BSc degree from Beni-Suef University, Egypt, in 2016. His research interests include Sentiment analysis and Machine learning.